\documentclass[conference]{IEEEtran}
\IEEEoverridecommandlockouts
\usepackage{cite}
\usepackage{amsmath,amssymb,amsfonts}
\usepackage{algorithm}
\usepackage{graphicx}
\usepackage{textcomp}
\usepackage{xcolor}
\usepackage{subfigure}
\usepackage{algpseudocode}
\usepackage{bm}
\usepackage{url}
\usepackage{balance}

\newcommand{\R}{\mathbb{R}}

\newcommand{\norm}[1]{\left\lVert #1 \right\rVert}

\let\oldeqref\eqref
\renewcommand{\eqref}[1]{eq.~\oldeqref{#1}}

\def\BibTeX{{\rm B\kern-.05em{\sc i\kern-.025em b}\kern-.08em
    T\kern-.1667em\lower.7ex\hbox{E}\kern-.125emX}}

\begin{document}

\title{BIND-USBL: Bounding IMU Navigation Drift using USBL in Heterogeneous ASV--AUV Teams
\thanks{\textsuperscript{1}Dept of Computer and Information Science, University of Konstanz,
Germany; \textsuperscript{2}Centre for the Advanced Study of Collective Behaviour,
University of Konstanz, Germany.
{\tt\{pranav.kedia, heiko.hamann\}@uni-konstanz.de}\par
\textsuperscript{3}Dept of Aerospace Engineering, Indian Institute of Science, Bangalore,
India. {\tt\{rajinimakam, vssuresh\}@iisc.ac.in}\\
Corresponding author: Pranav Kedia
}}

\author{Pranav Kedia\textsuperscript{1,2}, Rajini Makam\textsuperscript{3},
        Heiko Hamann\textsuperscript{1,2}, and Suresh Sundaram\textsuperscript{3}}

\maketitle

\begin{abstract}
Accurate and continuous localization of Autonomous Underwater Vehicles (AUVs) in
GPS-denied environments is a persistent challenge in marine robotics. In the absence
of external position fixes, AUVs rely on inertial dead-reckoning, which accumulates
unbounded drift due to sensor bias and noise. This paper presents BIND-USBL,
a cooperative localization framework in which a fleet of Autonomous Surface Vessels
(ASVs) equipped with Ultra-Short Baseline (USBL) acoustic positioning systems provides
intermittent fixes to bound AUV dead-reckoning error. The key insight is that
long-duration navigation failure is driven not by the accuracy of individual USBL
measurements, but by the temporal sparsity and geometric availability of those fixes. BIND-USBL combines a multi-ASV formation model linking survey scale and anchor placement to acoustic coverage, a conflict-graph-based TDMA uplink scheduler for shared-channel servicing, and delayed fusion of received USBL updates with drift-prone dead reckoning. The framework is evaluated in the HoloOcean simulator using heterogeneous ASV-AUV teams executing lawnmower coverage missions. The results show that localization performance is shaped by the interaction of survey scale, acoustic coverage, team composition, and ASV-formation geometry. Further, The spatial-reuse scheduler improves per-AUV fix delivery rate without violating the no-collision constraint, while maintaining low end-to-end fix latency.

\end{abstract}


\section{Introduction}

Marine robotics is undergoing a transformative shift toward heterogeneous multi-vehicle
systems in which \textit{Autonomous Surface Vessels} (ASVs) and
\textit{Autonomous Underwater Vehicles} (AUVs) operate in coordinated teams to
accomplish tasks that would be infeasible or prohibitively expensive with a single
platform~\cite{zhou2022survey}. Applications include wide-area seafloor mapping,
oceanographic sampling, search-and-rescue operations, infrastructure inspection,
and persistent environmental monitoring~\cite{makam2023}. In all such missions,
accurate and continuous localization of the underwater agents is paramount: an AUV
that does not know where it is cannot complete a survey, cannot rendezvous with a
recovery vehicle, and cannot safely avoid hazards.

\textit{Ultra-Short Baseline} (USBL) acoustic positioning is the most widely
deployed solution for operational AUV localization. A~USBL system uses a
multi-element transducer array mounted on a surface vessel to exchange acoustic
signals with an AUV-mounted transponder, estimating slant range via Two-Way Travel
Time (TWTT) and bearing via Time Difference of Arrival (TDOA)~\cite{Paull2014AUVNav}.
By combining these measurements with the GNSS-referenced position of the surface
vessel, an absolute AUV position fix is obtained. Compared to Long Baseline (LBL)
systems, which require pre-deployed infrastructure, and surfacing-based localization,
which disrupts missions, USBL provides a flexible and infrastructure-free alternative. 
USBL delivers frequent position updates over moderate ranges, thereby making it the de
facto standard for ship-based AUV navigation support~\cite{Bresciani2021ASV}.

An alternative to the conventional USBL paradigm is \textit{passive inverted
USBL} (piUSBL), in which the hydrophone array is mounted on the AUV rather
than the surface platform, and a single time-synchronized acoustic beacon
broadcasts a known waveform at a fixed rate~\cite{Rypkema2017,Rypkema2019piUSBL}.
Each AUV passively receives the signal and independently estimates its range
via one-way travel time (OWTT) and bearing via onboard beamforming, eliminating
the two-way interrogation cycle and enabling simultaneous localization of
arbitrarily many vehicles from a single beacon~\cite{Rypkema2022SynchClock}.
Recent work has extended piUSBL to fully submerged multi-agent teams using
bearing, elevation, and depth-difference (BEDD) measurements within a
factor-graph framework, removing the requirement for a surface-expressed
anchor entirely~\cite{Velasco2025BEDD}. Open-source implementations such as
Raspi2USBL~\cite{Huang2025Raspi2USBL} have further reduced the cost and
accessibility barrier, demonstrating sub-decimetre ranging accuracy and
bearing precision within $0.1^\circ$ at ranges up to $1.3$~km. While piUSBL
offers compelling scalability for large fleets, it requires clock
synchronization hardware (typically an oven-controlled crystal oscillator) on
every vehicle and delegates all signal processing to the AUV, increasing
onboard computational and power demands.

Replacing crewed support vessels with ASVs offers significant advantages in cost,
safety, and operational autonomy. Autonomous wind-powered surface vessel (AWSV) teams ~\cite{kedia2024sailswarm} are emerging as particularly attractive platforms for this role, their energy
autonomy removes the endurance constraints of battery- or
fuel-driven ASVs, enabling persistent multi-day deployments as
acoustic anchors without the need for recovery or
refuelling~\cite{SUN2025114918,an2021auto}. This makes AWSVs a
natural candidate for the surface support layer envisioned by
BIND-USBL, where sustained station-keeping at designated
formation vertices is the primary operational requirement. Early work by Bremnes~\textit{et al.}~\cite{Bremnes2022Hybrid}
formalized this concept for single-AUV support, introducing a hybrid controller that
maintains the ASV within a bounded operational region. This framework was extended to
multi-AUV scenarios using a switching control architecture with Extended Kalman Filter
(EKF) state estimation~\cite{Bremnes2024Switching}. Complementary efforts demonstrated
cooperative ASV--AUV systems with optimized acoustic protocols~\cite{Bresciani2021ASV},
and coordinated control strategies for search-and-rescue and trajectory tracking under
communication constraints~\cite{Wang2024SAR,Hu2025PTSMC}.

Despite these advantages, USBL-based localization is constrained by azimuth
estimation errors that grow proportionally with slant range~\cite{Costanzi2017,liu2024},
systematic biases from installation misalignment~\cite{ErrorUSBL2022}, and acoustic
propagation distortions from multipath and sound speed
variations~\cite{Bresciani2021ASV,liu2025}. In the intervals between USBL fixes,
AUVs rely on IMU dead-reckoning, whose error grows unboundedly due to sensor noise,
bias, and stochastic disturbances~\cite{DeadReckoningError2023}. Practical
cost-constrained deployments use low-cost MEMS IMUs that exhibit substantially
higher drift rates than high-grade inertial systems, and whose uncertainties often
exhibit non-Gaussian characteristics requiring robust estimation
frameworks~\cite{Li2022CoopNav}.

The accuracy of USBL localization is further influenced by the geometric relationship
between the surface platform and the submerged vehicle, a dependency characterized
via Fisher Information Matrix (FIM) analysis linking sensor geometry to
Cram\'{e}r--Rao bounds~\cite{Bo2020OptimalGeometry}. Active ASV repositioning to
maximize information gain can significantly improve tracking
accuracy~\cite{Turkoglu2026FIM}. Within the broader multi-AUV literature, cooperative
strategies spanning formation control, coverage, and distributed estimation have been
extensively studied~\cite{zhou2022survey,Das2016Formation,Okamoto2024Formation,Mao2026CBF,Hu2025IMMUKF}.

Despite these advances, a critical gap remains: how a \emph{fleet} of ASVs can be
systematically organized to provide bounded navigation support to multiple AUVs
through coordinated scheduling and positioning. Existing works predominantly address
single-ASV support for multiple AUVs~\cite{Bremnes2024Switching} or analyze USBL
measurement quality in isolation. The coupled interaction between scheduling policies,
acoustic channel contention, fix availability, and AUV drift dynamics has not been
rigorously treated.

This paper introduces BIND-USBL to address this gap. The key insight is that
long-term navigation failure is driven not by the accuracy of individual USBL
measurements, but by their \emph{temporal sparsity}: as inter-fix intervals increase,
dead-reckoning error grows unbounded. The objective of each ASV is to deliver
localization updates with sufficient frequency and geometric diversity to bound the
ensemble navigation error. The main contributions of BIND-USBL are:

(1)~A~multi-ASV formation geometry framework providing explicit acoustic coverage
      guarantees for the $N$-ASV, $N$-AUV setting via a corner-to-nearest-ASV bound
      and a closed-form minimum formation radius constraint.

(2)~An acoustic conflict graph formulation and greedy graph-colouring scheduling
      policy that enables spatial reuse of the shared acoustic uplink channel while
      preventing inter-AUV interference.

(3)~A~TDMA slot-timing model whose guard intervals scale with the acoustic crossing
      time of the survey area, ensuring protocol validity across operational scales.

We also incorporate multi-ASV USBL fix fusion via an MVLUE formulation reduce fused-fix standard deviation by$1/\sqrt{K}$. 
The framework is validated in the HoloOcean~\cite{holo} simulator with full vehicle dynamics, acoustic
propagation delay, packet loss, and communication scheduling.

The remainder of this paper is organized as follows. Section~\ref{BIND} formulates
the problem, presents vehicle dynamics, the USBL positioning model, multi-ASV
geometry and scheduling, the IMU drift model, and the complementary filter fusion.
Section~\ref{sim} presents simulation results. Section~\ref{con} concludes with
findings and future directions.

\section{BIND-USBL} \label{BIND}

\subsection{Problem Formulation}

Consider a multi-vehicle marine system comprising $N_\mathrm{AUV}$ AUVs and
$N_\mathrm{ASV}$ ASVs operating in a square survey area
$\Omega = [-L/2,\,L/2]^2 \subset \R^2$. GPS is unavailable underwater, so each
AUV $i$ propagates its position estimate via strapdown inertial dead-reckoning.
Without external correction, this estimate diverges from the true position due to sensor bias $\mathbf{b}_i$ and process noise, with error growing unboundedly over time. Each ASV is equipped with a USBL transducer array and GNSS, enabling it to compute absolute position fixes for AUVs within acoustic range. ASV's primary task is to deliver drift-bounding position fixes.

The system operates a two-band acoustic architecture: a short-range high-frequency (HF) uplink channel on which AUVs ping requesting fixes,
and a longer-range medium-frequency (MF) downlink channel on which ASVs broadcast the accumulated fix payload to AUVs. HF scheduling is managed by a TDMA protocol, whereas MF uses an event-triggered priority assignment.

The objective of BIND-USBL is to characterize how navigation accuracy depends on team composition and surface-formation geometry in heterogeneous ASV–AUV systems. For a given combination of AUVs and ASVs, the framework evaluates how the acoustic servicing rate, anchor placement, and multi-vehicle contention affect the achievable position error in practice. In this way, formation design is treated not as a mechanism for guaranteeing a prescribed bound, but as a system-level factor that shapes the trade-off between fix availability, localization quality, and resulting navigation accuracy.

\subsection{Vehicle Dynamics}

AUVs are modeled with 6-DOF nonlinear dynamics~\cite{fossen2011,makam2023}.
In simulation, the navigation filter operates on the 2D horizontal position
$\hat{\mathbf{p}}_i \in \R^2$; depth is handled independently via a pressure
sensor.

The ASV is modeled as a 3-DOF surface craft (surge, sway, yaw)~\cite{fossen2011}.
These dynamics are realized in
HoloOcean~\cite{holo}. A station-keeping controller holds the ASV near the survey
centroid, $\mathbf{p}_{T,\mathrm{ASV}} \approx \operatorname{centroid}(\Omega)$, through continuous thruster operation while preserving
favorable USBL geometry.

\subsection{USBL Navigation Principle \cite{max_phd}}

Each ASV estimates the relative position of an AUV using acoustic ranging and
direction finding. The slant range is obtained from the round-trip travel time (RTT):
\begin{equation}
  r = \frac{c\,\tau_\mathrm{RTT}}{2},
  \label{eq:rtt}
\end{equation}
where $c = 1500$~m/s is the nominal speed of sound.

The direction-of-arrival (DOA) is computed from the time-difference-of-arrival (TDOA)
measurements across the five-element hydrophone array. 
%
%
The ASV calculates the absolute world-frame fix of each AUV by combining the GNSS position
$\mathbf{p}_j^\mathrm{ASV}$ with the range $d_{ij} = c\,t_\mathrm{RTT}/2$ and
the TDOA-derived direction angles $(\theta, \phi)$ (azimuth, elevation):
\begin{equation}
  \mathbf{z}_{ij,k}
    = \mathbf{p}_{j,k}^\mathrm{ASV}
      + d_{ij}
      \begin{bmatrix}
        \cos\phi\cos\theta \\ \cos\phi\sin\theta \\ \sin\phi
      \end{bmatrix}
      + \bm{\epsilon}_\mathrm{USBL},
  \label{eq:usbl_fix}
\end{equation}
where, $i \in [1, N_{AUV}]$,  $j \in [1, N_{ASV}]$, $\bm{\epsilon}_\mathrm{USBL} \sim \mathcal{N}(\mathbf{0}, \Sigma_\mathrm{USBL})$
encodes RTT quantisation, TDOA noise, and AHRS heading error. When more than one ASV simultaneously hears AUV~$i$, the resulting multi-ASV fused fix is the
minimum-variance linear unbiased estimator (MVLUE).
This shows that deploying
additional ASVs improves not only fix \emph{availability} but also fix \emph{quality}.

Acoustic fix loss due to attenuation, multipath, and channel contention is modelled
probabilistically. The range-dependent loss probability follows~\cite{max_phd}:
\begin{equation}
  P_\mathrm{loss}(r)
    = a\,e^{b\,\tilde{r}} + c_0\,e^{d\,\tilde{r}},
  \quad \tilde{r} = \min(r,\,800\,\mathrm{m}),
  \label{eq:fix_loss}
\end{equation}
with coefficients $\{a, b, c_0, d\} =
\{-6.070,\;2.12\!\times\!10^{-3},\;5.987,\;2.25\!\times\!10^{-3}\}$.
In multi-AUV operations, simultaneous uplink pings elevate the collision probability:
\begin{equation}
  P_\mathrm{loss}^\mathrm{total}(r)
    = \min\!\bigl(P_\mathrm{loss}(r)
      + (N_\mathrm{AUV}-1)\,P_\mathrm{col},\;0.999\bigr),
  \label{eq:fix_loss_total}
\end{equation}
where $P_\mathrm{col} = 0.05$ per additional AUV. A fix is declared lost if
$u \sim \mathcal{U}(0,1)$ satisfies $u < P_\mathrm{loss}^\mathrm{total}$, or if
$r > r_\mathrm{max}$. This model from \cite{max_phd} is applied only in the range where it is physically valid.

\subsection{Multi-ASV Formation Geometry, Coverage and Conflict Graph}
\label{sec:formation}

The $N_\mathrm{ASV}$ ASVs are placed at the vertices of a regular $N$-gon centred
at the survey origin $\mathbf{0} \in \R^2$. The formation radius is
$R_f = R_\mathrm{HF} + \Delta_b$, where $\Delta_b \geq 0$ is a clearance buffer.
The Cartesian position of $ASV_j$ $(j = 0,\ldots,N_\mathrm{ASV}{-}1)$ in the NED
plane is:
\begin{equation}
  \mathbf{p}_j = R_f
  \begin{bmatrix}
    \cos \left(\alpha_0 + \tfrac{2\pi j}{N_\mathrm{ASV}}\right) \\[3pt]
    \sin\left(\alpha_0 + \tfrac{2\pi j}{N_\mathrm{ASV}}\right)
  \end{bmatrix},
  \quad R_f = R_\mathrm{HF} + \Delta_b,
  \label{eq:formation}
\end{equation}
with formation angle $\alpha_0$ (default $0$). For $N_\mathrm{ASV} = 1$, the
formation degenerates to $\mathbf{p}_0 = \mathbf{0}$.

The maximum distance from any point in $\Omega$ to the nearest ASV is bounded by
the corner-to-nearest-ASV distance:
\begin{equation}
  d_\mathrm{corner}
    = \min_{j}\norm{\mathbf{q}_c - \mathbf{p}_j},
  \quad \mathbf{q}_c \in \bigl\{\pm\tfrac{L}{2}\bigr\}^2.
  \label{eq:corner_dist}
\end{equation}
Full acoustic coverage requires $d_\mathrm{corner} \leq R_\mathrm{HF}$, giving
the necessary formation radius constraint:
\begin{equation}
  R_f \;\geq\; \frac{L}{2}
    - \sqrt{R_\mathrm{HF}^2 - \left(\frac{L}{2}\right)^{\!2}},
  \label{eq:coverage}
\end{equation}
valid when $R_\mathrm{HF} \geq L/2$. For the default parameters
($L = 60$~m, $R_\mathrm{HF} = 50$~m), the right-hand side of~\eqref{eq:coverage}
equals $-10$~m, so any positive formation radius satisfies full coverage, confirming
that the two-band acoustic architecture provides adequate uplink coverage at nominal
scale.


The HF uplink is a shared acoustic channel: two AUVs pinging simultaneously cause
a collision if any ASV lies within range of both. Two AUVs $i$ and $j$ are in
\emph{acoustic conflict} if:
\begin{eqnarray}
  \mathrm{AC}(i,j)
    \;\Leftrightarrow\;
    \exists\,k \in [1,N_\mathrm{ASV}] :
      d(\mathbf{p}_i, \mathbf{p}_{\mathrm{ASV},k}) \leq R_\mathrm{HF}
      \;\wedge\; \nonumber \\
      d(\mathbf{p}_j, \mathbf{p}_{\mathrm{ASV},k}) \leq R_\mathrm{HF},
  \label{eq:conflict_pred}
\end{eqnarray}
where $\mathbf{p}_{\mathrm{ASV},k}$ is the position of the $k$-th ASV. 
That means two AUVs are in acoustic conflict whenever at least one ASV can hear both of them, so simultaneous uplink transmissions could collide at that ASV. 
The conflict graph $G = (V, E)$ has vertex set $V = \{\mathrm{AUV}_1, \ldots, \mathrm{AUV}_{N_\mathrm{AUV}}\}$
and edge set $E = \{(i,j) : \mathrm{AC}(i,j)\}$. Since AUV positions change
continuously, $G$ is recomputed at the end of every downlink message.

A proper $k$-colouring \cite{diestel2017graph} assigns colour $c(i) \in \{0,\ldots,k-1\}$ to each vertex
such that adjacent vertices have different colours. AUVs sharing a colour may ping
simultaneously without collision. The greedy algorithm assigns:
\begin{equation}
  c(i) = \min\bigl\{c \in \mathbb{N}_0 :
    c \notin \{c(j) : j \in \mathcal{N}(i),\;c(j) \neq -1\}\bigr\},
  \label{eq:greedy_colour}
\end{equation}
where $\mathcal{N}(i) = \{j : (i,j) \in E\}$. The resulting colour classes
$\mathcal{G}_0,\ldots,\mathcal{G}_{k-1}$ partition $V$ and define the TDMA ping
groups. By Brooks' theorem, $k \leq \Delta(G) + 1$, where $\Delta(G)$ is the
maximum vertex degree, bounding the number of slots in the worst case.

\subsection{TDMA Acoustic Protocol}
\label{sec:tdma}

A TDMA slot must accommodate: (i)~ping transmission $T_\mathrm{P}$,
(ii)~worst-case one-way travel time (OWTT) $\tau_\mathrm{OT,max}$, and
(iii)~a guard time $T_\mathrm{G}$ against multipath echo arrivals. All timing
constraints are expressed in units of the \emph{acoustic crossing time}:
\begin{equation}
  t_\mathrm{C} = \frac{L}{c} = \frac{L}{1500}\;\mathrm{[s]},
  \label{eq:t_cross}
\end{equation}
which scales guard intervals (idle times inserted between transmissions) with the survey size and ensures the protocol remains physically valid as $L$ changes. The uplink slot duration for a ping group is:
\begin{equation}
  T_\mathrm{UL} = \max\!\left\{
    T_\mathrm{P} + \tau_\mathrm{OT,max} + T_\mathrm{G},\;
    T_\mathrm{UL,min}
  \right\},
  \label{eq:UL_slot}
\end{equation}
where $T_\mathrm{P} = 10$~ms (fixed hardware), and:
\begin{align}
  T_\mathrm{G}        &= 0.5\,t_\mathrm{C},  \label{eq:guard} \\
  T_\mathrm{UL,min}   &= 2.5\,t_\mathrm{C}.  \label{eq:UL_min}
\end{align}

The guard factor $0.5$ bounds multipath delay spread within the survey area; the
minimum slot factor $2.5$ provides a margin for inter-agent timing jitter.
Let $k_\mathrm{S}^{g}$ denote the start tick of group~$g$'s uplink slot and
$f_\mathrm{T} = 30$\,Hz the simulation tick rate
($\Delta t = 1/f_\mathrm{T}$). The earliest tick at which group $g{+}1$ may
begin transmitting is:
\begin{equation}
  k_\mathrm{S}^{g+1}
    = k_\mathrm{S}^{g}
      + \bigl\lceil T_\mathrm{UL}^{g}\,f_\mathrm{T} \bigr\rceil,
  \label{eq:UL_tick}
\end{equation}
where the ceiling converts the continuous-time slot duration into an integer
tick count. 

After all colour groups in the current round have pinged, one ASV broadcasts on
the shared MF downlink channel. The broadcast payload comprises an $N_\mathrm{hdr}
= 8$\,byte header followed by $K_\mathrm{fix}$ individual fix records of
$B_\mathrm{fix} = 16$\,bytes each, where $K_\mathrm{fix}$ is the number of
fixes accumulated during the uplink round:
\begin{equation}
  N_\mathrm{bytes}
    = N_\mathrm{hdr} + K_\mathrm{fix}\,B_\mathrm{fix}
    \;\mathrm{[bytes]}.
  \label{eq:payload}
\end{equation}
The transmission duration (with protocol overhead factor $\Lambda = 2.0$) is:
\begin{equation}
  T_\mathrm{tx}
    = \frac{N_\mathrm{bytes} \cdot 8 \cdot \Lambda}{R_\mathrm{DL}},
  \qquad R_\mathrm{DL} = 2000\;\mathrm{bps}.
  \label{eq:Ttx}
\end{equation}

The total MF slot duration is:
\begin{align}
     T_\mathrm{DL}   & = \max \left\{T_\mathrm{tx} + T_\mathrm{DL,G},\;T_\mathrm{DL,min}\right\}, \label{eq:DL_slot}
\end{align}
with,
\begin{eqnarray}
     T_\mathrm{DL,G}    &= 1.25\,t_\mathrm{C}, \label{eq:DL_guard} \\
  T_\mathrm{DL,min}  &= 10.0\,t_\mathrm{C}. \label{eq:DL_min} 
\end{eqnarray}

The large minimum downlink slot ($\alpha_\mathrm{DL,min} = 10.0$) is chosen
conservatively to accommodate variable acoustic propagation delay across the full
MF range and to allow reliable decoding at low SNR.

A fix broadcast at tick $k_\mathrm{B}$ by ASV $j$ reaches AUV $i$ at:
\begin{equation}
  k_\mathrm{D}
    = k_\mathrm{B}
      + \left\lceil\!\left(T_\mathrm{tx} + \frac{d_{ij}}{c}\right)\!f_\mathrm{T}
        \right\rceil,
  \label{eq:deliver_tick}
\end{equation}
$d_{ij} = \norm{\mathbf{p}_{\mathrm{ASV},j}^{xy} - \mathbf{p}_{\mathrm{AUV},i}^{xy}}$
is the horizontal separation at broadcast time. If $d_{ij} > R_\mathrm{MF}$, the
fix is not delivered. The end-to-end latency $L_\mathrm{E2E}$ from HF ping to fix delivery is:
\begin{equation}
  L_\mathrm{E2E}
    = \bigl(k_\mathrm{D} - k_\mathrm{P}\bigr)\,\Delta T,
  \label{eq:E2E_latency}
\end{equation}
where $k_\mathrm{P}$ is the HF ping tick and $\Delta T = 1/f_\mathrm{T}$. 


\subsection{Strapdown IMU Drift Model}
\label{sec:imu}

In the absence of acoustic fixes, each AUV $i$ propagates its estimated 2D
horizontal position via strapdown dead-reckoning corrupted by a constant velocity
bias $\mathbf{b}_i \in \R^2$ and additive noise:
\begin{equation}
  \hat{\mathbf{p}}_i(k{+}1)
    = \hat{\mathbf{p}}_i(k)
      + \mathbf{R}(\psi_k)\!\left(
          \mathbf{v}_k^\mathrm{true}\Delta t
          + \mathbf{b}_i\Delta t
          + \bm{\eta}_k\sqrt{\Delta t}
        \right),
  \label{eq:dead_reckoning}
\end{equation}
where $\mathbf{R}(\psi) \in SO(2)$ is the yaw rotation matrix and
$\bm{\eta}_k \sim \mathcal{N}(\mathbf{0},\sigma^2\mathbf{I})$.
Note that $\mathbf{v}_k^\mathrm{true}$ is the ground-truth velocity provided by
the HoloOcean simulator; in a real deployment this would be replaced by an
accelerometer-derived velocity estimate, adding a further source of error.

The simulation parameters are $\mathbf{b} = [0.06, 0.06]^\top$~m/s and
$\sigma = 0.027$~m/$\sqrt{\mathrm{s}}$. The bias-induced position error grows
linearly in continuous time:
\begin{equation}
  \norm{\mathbf{e}_\mathrm{DR}(t)}
    \approx \tfrac{1}{2}\norm{\mathbf{b}_i}\,t,
  \label{eq:drift_growth}
\end{equation}
and the total RMS error envelope (bias ramp plus random walk) is:
\begin{equation}
  \sigma_\mathrm{DR}(t)
    \approx \sqrt{\tfrac{1}{4}\norm{\mathbf{b}_i}^2 t^2
                  + \sigma^2 t}.
  \label{eq:error_envelope}
\end{equation}

Depth is measured independently by a pressure sensor with Gaussian noise:
$\tilde{z}(k) = z(k) + \varepsilon_z(k)$, $\varepsilon_z \sim \mathcal{N}(0,\sigma_z^2)$,
$\sigma_z = 0.05$~m. The vertical state component is replaced in both the IMU and
fused estimates at every tick, eliminating depth-axis drift:
\begin{equation}
  \hat{x}_{\mathrm{IMU},z}(k) \leftarrow \tilde{z}(k),
  \qquad
  \hat{x}_{\mathrm{fused},z}(k) \leftarrow \tilde{z}(k).
  \label{eq:depth_replace}
\end{equation}


Upon delivery of fix $\hat{\mathbf{z}}_{i,k}$, the horizontal position estimate
is updated via a two-step predict--correct scheme with gain $\gamma \in (0,1]$:
\begin{align}
  \hat{\mathbf{p}}_i^-(k)
    &= \hat{\mathbf{p}}_i(k{-}1)
       + \mathbf{R}(\psi_k)\bigl(\mathbf{v}_k^\mathrm{true}
         + \mathbf{b}_i\bigr)\Delta t,
  \label{eq:predict}\\[2pt]
  \hat{\mathbf{p}}_i(k)
    &= \hat{\mathbf{p}}_i^-(k)
       + \gamma\bigl[\hat{\mathbf{z}}_{i,k}
                     - \hat{\mathbf{p}}_i^-(k)\bigr].
  \label{eq:update}
\end{align}
Under the EKF interpretation with measurement matrix $\mathbf{H} = \mathbf{I}$,
the posterior covariance satisfies $\mathbf{P}^+ = (1-\gamma)\mathbf{P}^-$.
The fixed gain $\gamma = 0.90$ is used throughout BIND-USBL, applying 90\% of
the innovation as a position correction at each fix delivery. Fixes from the ASV
are stored in a per-AUV min-heap priority queue keyed on delivery tick $k_\mathrm{D}$,
ensuring causal application and consistent temporal ordering even under variable
acoustic delay.

\section{Simulation Results} \label{sim}

We evaluate BIND-USBL in \textit{HoloOcean}~\cite{holo} through a set of GPS-denied lawnmower coverage missions involving heterogeneous ASV--AUV teams. We begin with an ideal single-ASV baseline, then examine representative failure and recovery cases in larger survey areas, and finally report the full
configuration sweep across ASV count, formation geometry, and survey scale.

\subsection{Experimental Setup}
 
All experiments are conducted through simulations in \textit{HoloOcean}. AUVs execute Dubins-smooth boustrophedon lawnmower missions at a fixed depth of $10$~m; the survey area $\Omega = [-L/2,\,L/2]^2$ is partitioned into equal-width horizontal strips, one per AUV. ASVs are deployed in a stationary regular-$N$-gon formation centred at the survey origin and held in station-keeping throughout each mission.
 
The two-band acoustic protocol described in Section~\ref{BIND} is active throughout.
HF uplink parameters are $R_\mathrm{HF} = 50$~m and $R_\mathrm{MF} = 100$~m. These ranges are shorter than
typical acoustic and USBL systems, where HF channels may
reach several hundred metres and MF links extend to
kilometre-scale distances; however, the $2{:}1$ ratio between
downlink and uplink range is preserved, maintaining the
architectural relationship between the two bands. The
reduced absolute values allow the coverage-failure regime
to be reached within the simulator's spatial constraints,
making the interaction between survey scale, formation
geometry, and acoustic coverage experimentally accessible
at $L \leq 140$~m.

USBL fixes are generated using the noise and outage model of~\cite{max_phd}
(RTT quantisation, TDOA noise, AHRS heading error, range-dependent and team-size based packet loss).
The IMU dead-reckoning bias is $\mathbf{b} = [0.06,\,0.06]^\top$~m/s, step noise
$\sigma = 0.027$~m/$\sqrt{\mathrm{s}}$, and depth noise $\sigma_z = 0.05$~m.
The complementary filter gain is fixed at $\gamma = 0.90$. All simulations run at
$30$~Hz; unless stated otherwise, results are averaged over $20$ independent random
seeds and all applicable formation angles.
 
Three survey area side lengths are tested: $L \in \{60,\,100,\,140\}$~m. The primary
performance metrics are mean cross-track error (CTE) in metres, per-AUV HF coverage
fraction (fraction of ping attempts heard by at least one ASV), cumulative fix count,
and end-to-end acoustic latency. Results are structured to trace the progression from
a well-conditioned baseline to a coverage-limited failure regime, and then to show
how BIND-USBL's multi-ASV formation and scheduling framework resolves each failure
mode.

\subsection{Baseline Validation in a Single-ASV Coverage Mission}
\label{sec:baseline}

We first establish a reference operating point in which a single station-kept ASV
supports four AUVs over a $60\times 60$~m survey for $T = 300$~s. At this scale,
the entire survey area lies within $R_\mathrm{HF} = 100$~m of the origin: every AUV
is audible to the ASV throughout the mission. The acoustic conditions are therefore
ideal, and this experiment isolates the behaviour of the TDMA scheduler and the
complementary filter without coverage-induced degradation.
 
\begin{figure}[htbp]
\centering
\includegraphics[width=\columnwidth]{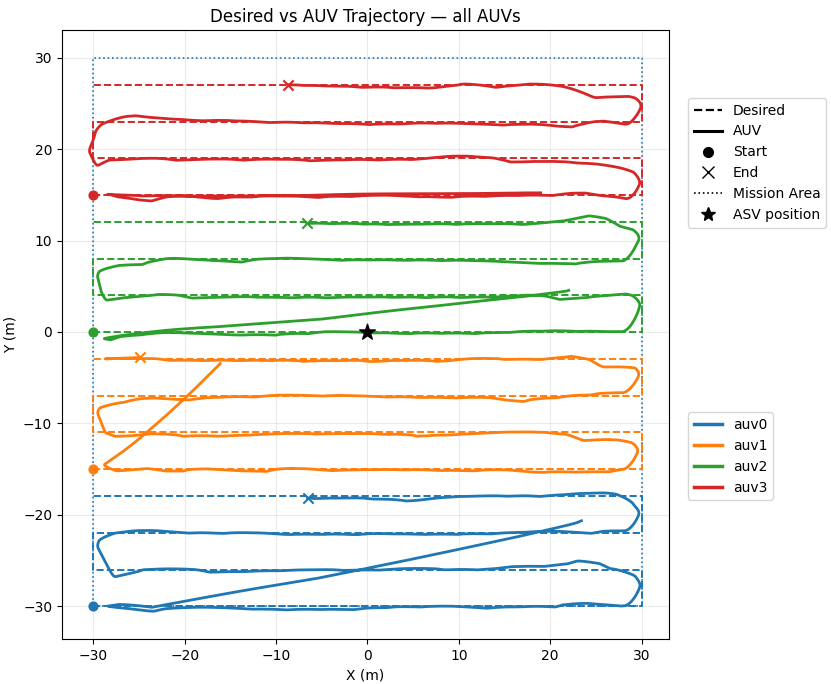}
\caption{Multi-AUV boustrophedon coverage at $L = 60$~m. The single ASV
(star) is station-kept at the survey centroid; all four lawnmower strips
fall within its HF uplink range = 100~m.}
\label{fig:desired_vs_actual_gt}
\end{figure}

Fig.~\ref{fig:desired_vs_actual_gt} shows the spatial layout of the mission. The
IMU-only trajectory exhibits the characteristic piecewise quadratic drift predicted
by \eqref{eq:drift_growth}; the fused IMU+USBL trajectory tracks the planned path
closely, demonstrating that a steady supply of USBL fixes is sufficient to bound
inertial drift.
 
Across five independent runs, the TDMA scheduler delivers an aggregate applied
fix rate of $0.994$~Hz (in average $298.2$ fixes over $300$~s), with a per-AUV rate of
$0.248$~Hz and a mean inter-fix interval of $4.02$~s. Fix allocation is
near-uniform: $24.3\%$, $25.9\%$, $24.4\%$, and $25.4\%$ across
AUV$_0$--AUV$_3$. End-to-end acoustic latency is $51$~ms (mean) with
$p_{95} = 67$~ms. With the simulation bias $\|\mathbf{b}\| \approx 0.085$~m/s and
mean inter-fix interval $T_\mathrm{IFI} = 4.02$~s, the maximum bias-induced drift
per fix interval from~\eqref{eq:drift_growth} is
$\tfrac{1}{2}\times 0.085\times 4.02^2 \approx 0.69$~m, consistent with the
sub-metre CTE observed in practice. This baseline confirms that the TDMA protocol
correctly services the fleet at the maximum achievable rate given a single shared
channel, and that the complementary filter successfully bounds drift to operationally
acceptable levels.
 
The baseline, however, relies critically on full acoustic coverage. The next
experiment investigates the consequences when the survey area extends beyond the single-ASV HF footprint.

 
\subsection{Coverage Failure and Recovery}
\label{sec:case_study}

We scale the survey to $L = 140$~m with $3$~AUVs and $1$~ASV, keeping all other
parameters fixed. At this scale the corner-to-ASV distance is
$\|\mathbf{q}_c - \mathbf{0}\| = L\sqrt{2}/2 \approx 99$~m, nearly twice the HF
range of $50$~m. The single centroid-stationed ASV therefore cannot provide HF
coverage to the survey extremes, and some AUVs will inevitably operate outside its
acoustic footprint for extended periods.
 
\begin{figure}[tbp]
  \centering
  \includegraphics[width=\columnwidth]{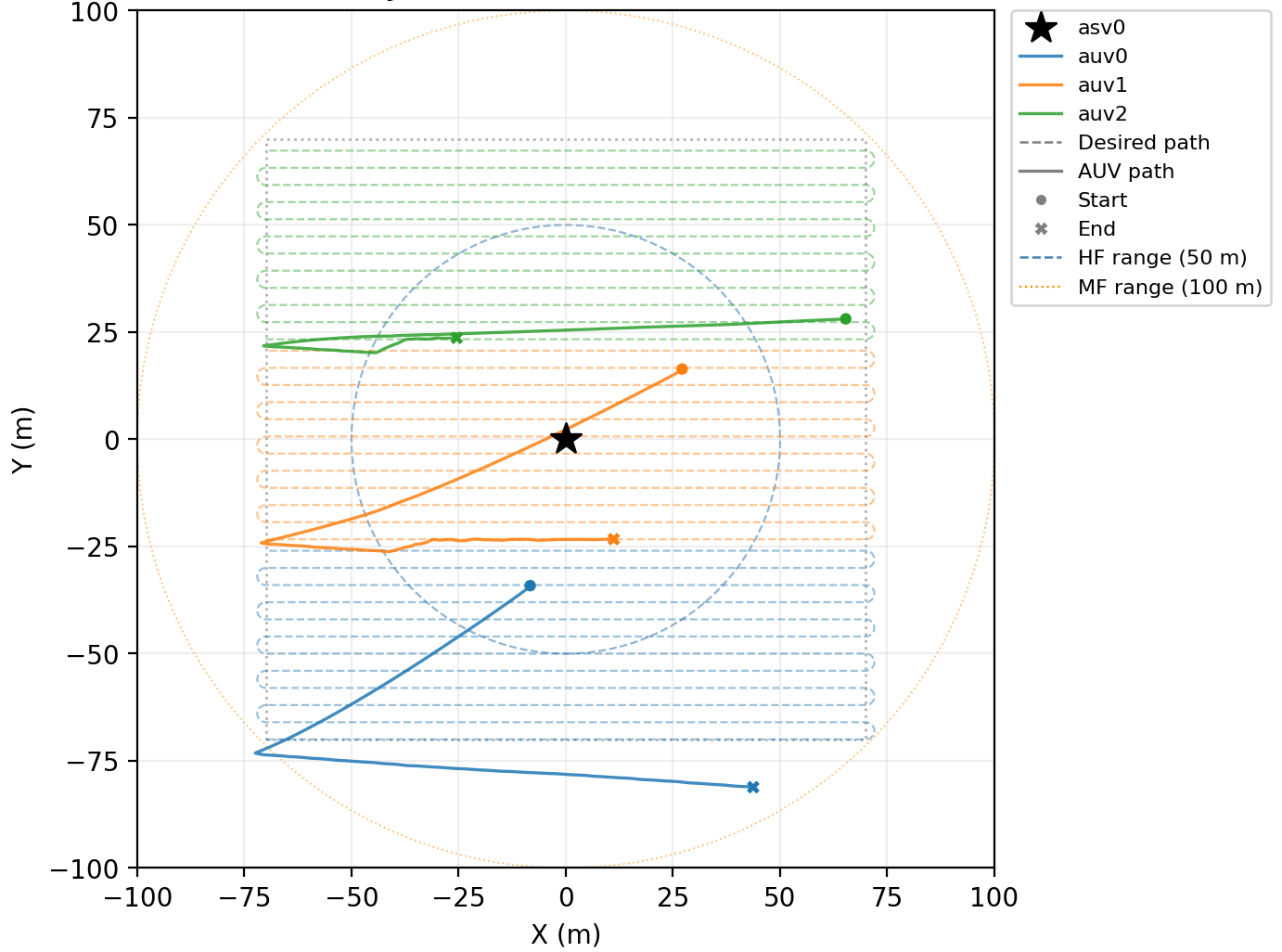}
  \caption{Trajectory overview: $L = 140$~m, $3$~AUVs, $1$~ASV
  (star, $\alpha_0 = 0^\circ$). The dashed blue circle marks the HF uplink
  range, and the dotted orange circle marks the MF downlink range. The
  upper lawnmower strip lies almost entirely outside the HF footprint.}
  \label{fig:case_fail_xy}
\end{figure}

 
Fig.~\ref{fig:case_fail_xy} shows the trajectory overview for Single-ASV configuration
($\alpha_0 = 0^\circ$). The per-AUV breakdown reveals extreme servicing inequality
driven entirely by acoustic geometry. AUV$_0$, whose lawnmower strip occupies the
$y$-extreme of the survey area, exits the ASV's HF footprint almost immediately and
never re-enters it for meaningful durations. It achieves only $9\%$ HF coverage,
receiving just $19$ fixes over the full $300$~s mission.
 
\begin{figure*}[htbp]
  \centering
  \includegraphics[width=1.5\columnwidth]{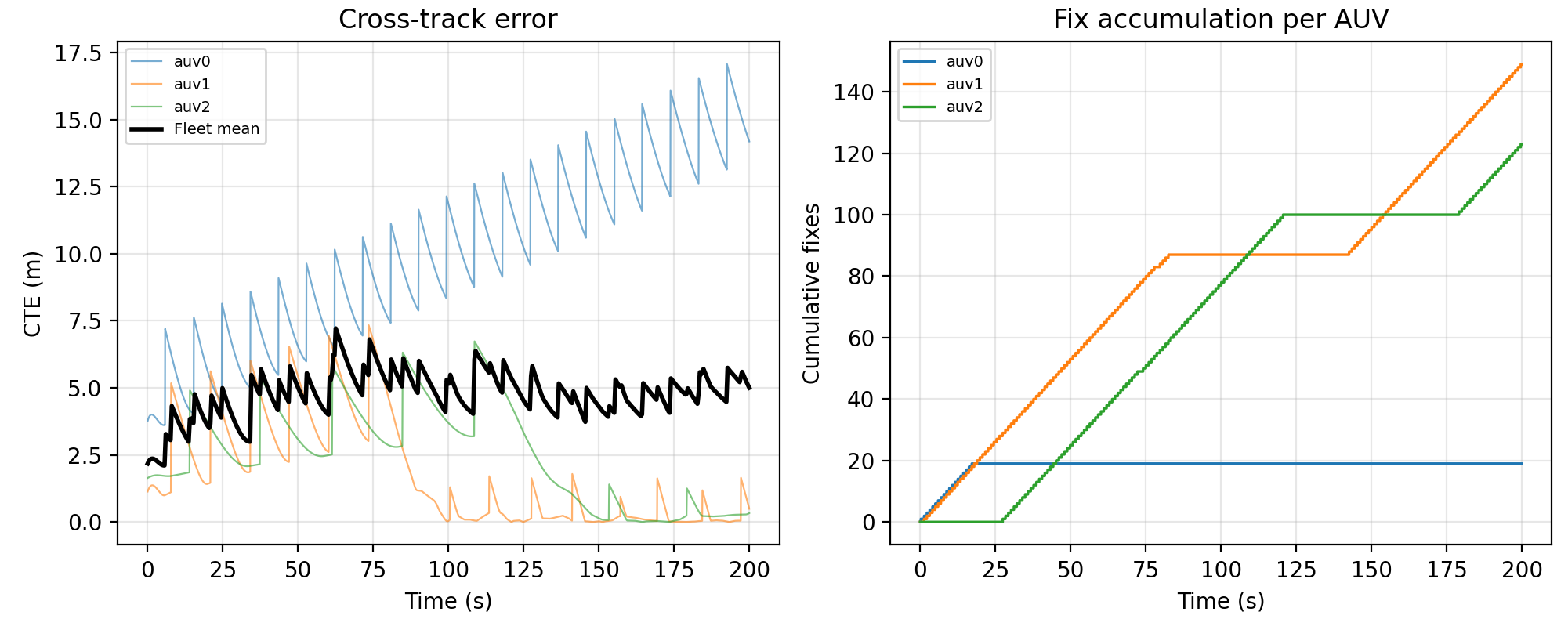}
  \caption{Fleet metrics for the single-ASV failure case ($L = 140$~m,
  $3$~AUVs, $1$~ASV, $\alpha_0 = 0^\circ$). Left: cross-track error over
  time; AUV$_0$ diverges unboundedly while AUV$_1$--AUV$_2$ are bounded.
  Right: cumulative fix count; AUV$_0$'s count flatlines after the first
  few seconds, characteristic of coverage-limited starvation.}
  \label{fig:case_fail_metrics}
\end{figure*}

The consequence is clear in Fig.~\ref{fig:case_fail_metrics}: AUV$_0$'s cumulative
fix count flatlines early in the mission while AUV$_1$ and AUV$_2$ accumulate fixes
at a near-linear rate. Deprived of corrections, AUV$_0$ follows the unbounded
quadratic drift trajectory of~\eqref{eq:drift_growth}, accumulating a mean CTE of
$10.0$~m. AUV$_1$ and AUV$_2$, operating closer to the centroid with $70\%$ and
$58\%$ coverage respectively, maintain CTE of $1.79$~m and $2.52$~m. All three AUVs
travel comparable distances ($182$--$192$~m), confirming that the performance gap is
entirely acoustic-geometric in origin (Table~\ref{tab:case_study}).
 
This failure mode is binary, not gradual: a~vehicle either remains within the ASV's
HF footprint and receives adequate fixes, or it exits the footprint and drifts
uncorrected. The TDMA scheduler cannot compensate for this failure mode because no valid USBL measurements are available to be delivered to AUV$_0$. The root cause is architectural:
a single surface anchor with a fixed HF range cannot provide continuous
coverage of a survey whose diagonal exceeds that range. 
\begin{figure*}
      \centering
  \includegraphics[width=1.4\columnwidth]{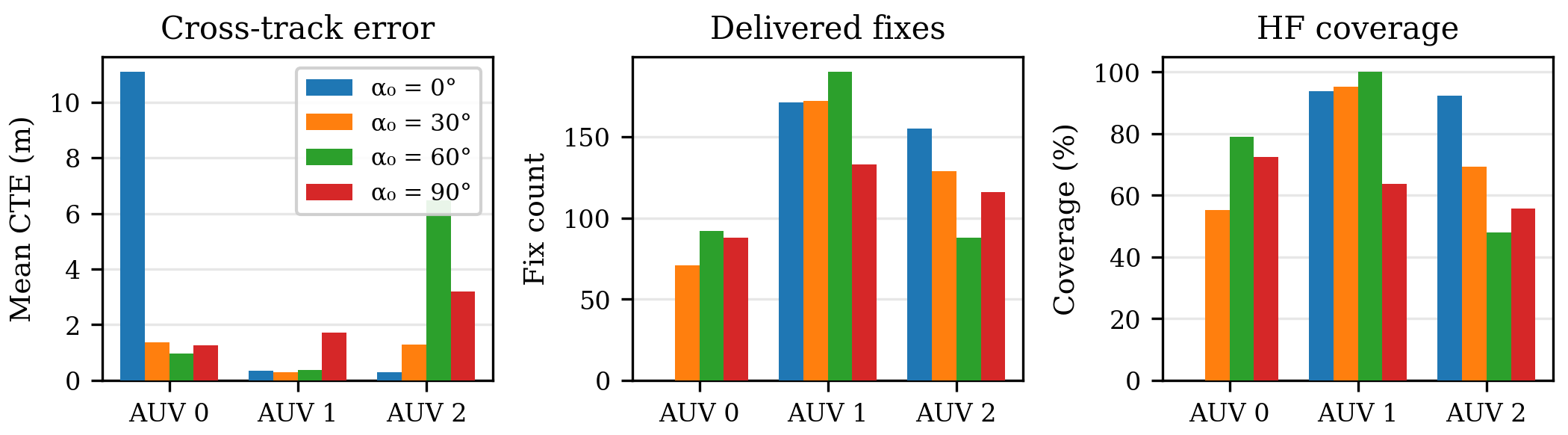}
  \caption{Per-AUV CTE, fix count, and HF coverage for $L = 140$~m,
  $3$~AUVs, $2$~ASVs across formation angles $\alpha_0 \in \{0^\circ,
  30^\circ, 60^\circ, 90^\circ\}$ (single representative run per angle).
  At $\alpha_0 = 0^\circ$ the two-ASV pair is aligned along the $x$-axis,
  leaving $y$-extreme AUVs unserved. A $30^\circ$ rotation yields an
  $8\times$ CTE reduction for AUV$_0$ with no change in hardware.}
  \label{fig:case_angle}
\end{figure*}
 
 
BIND-USBL addresses this failure mode directly through its $N$-gon formation geometry. Deploying three ASVs in an equilateral triangle
formation ($\alpha_0 = 0^\circ$) at radius $R_f = R_\mathrm{HF} = 50$~m distributes
acoustic coverage across the survey area from multiple directions, ensuring that each
lawnmower strip lies within HF range of at least one ASV at all times.

This reverses the previously observed failure mode.
AUV$_0$, which previously
received only $19$ fixes with $9\%$ coverage, now achieves $79\%$ coverage, $92$ delivered fixes, and a mean CTE of $0.56$~m. 
AUV$_1$ and AUV$_2$ reach $100\%$ coverage with CTE
below $0.35$~m (Table~\ref{tab:case_study}). The multi-ASV deployment substantially improved coverage from 9\% to 79\%, demonstrating partial geometric recovery. Note that the formal guarantee of \eqref{eq:coverage}  applies only when $R_HF \geq L/2$. The fix-rate improvement translates directly to navigation
accuracy through the bounded inter-fix interval of~\eqref{eq:drift_growth},
demonstrating the core BIND-USBL design principle.
 
\begin{table}[tbp]
  \centering
  \caption{Per-AUV metrics for the coverage-failure and recovery cases
  ($L = 140$~m, $3$~AUVs, single representative run).}
  \label{tab:case_study}
  \renewcommand{\arraystretch}{1.0}
  \begin{tabular}{l l r r r r}
    \hline
    \textbf{Config} & \textbf{AUV} & \textbf{Fixes} &
    \textbf{Cov.\,(\%)} & \textbf{CTE\,(m)} & \textbf{Dist.\,(m)} \\
    \hline
    1 ASV, $0^\circ$ & AUV$_0$ &  19 &   9 & 10.02 & 192 \\
                      & AUV$_1$ & 149 &  70 &  1.79 & 190 \\
                      & AUV$_2$ & 123 &  58 &  2.52 & 182 \\
    \hline
    3 ASVs, $0^\circ$ & AUV$_0$ &  92 &  79 &  0.56 & 192 \\
                       & AUV$_1$ & 189 & 100 &  0.35 & 190 \\
                       & AUV$_2$ & 185 & 100 &  0.35 & 182 \\
    \hline
  \end{tabular}
\end{table}
 
 
The coverage-failure mode is not unique to single-ASV deployments. For two ASVs,
the formation angle $\alpha_0$ determines whether the pair's footprints complement
each other or leave symmetric gaps. Fig.~\ref{fig:case_angle} shows the per-AUV
breakdown for two ASVs at $L = 140$~m across four formation angles.

At $\alpha_0 = 0^\circ$, the two ASVs are aligned along the $x$-axis, replicating
the same coverage shadow at the $y$-extremes. AUV$_0$ receives \emph{zero} fixes
and drifts to $11.1$~m CTE. Rotating the formation to $\alpha_0 = 30^\circ$ breaks
the alignment symmetry: AUV$_0$ recovers to $55\%$ coverage ($71$ fixes) and
$1.37$~m CTE---an $8\times$ improvement from a $30^\circ$ rotation alone, with
identical hardware. This confirms that for two-ASV deployments, $\alpha_0$ is a
first-order design variable that can determine mission success or failure. The
BIND-USBL formation model provides the analytical
basis for choosing $\alpha_0$ to maximise the worst-case corner coverage distance.

 
\subsection{Configuration Sweep: ASV Count and Survey Scale}
\label{sec:sweep}

Having established the failure mechanism and recovery pathway, we report the full
parameter sweep across ASV count (up to three), formation angle
($0^\circ$--$90^\circ$), survey side length ($L \in \{60, 100, 140\}$~m), and AUV
count ($3$--$10$). Results are averaged over $20$ independent seeds and all four
formation angles; the distance-normalised protocol holds total team travel
approximately constant across team sizes to isolate the effect of composition from
mission duration.
\begin{figure}[htbp]
  \centering
  \includegraphics[width=\columnwidth]{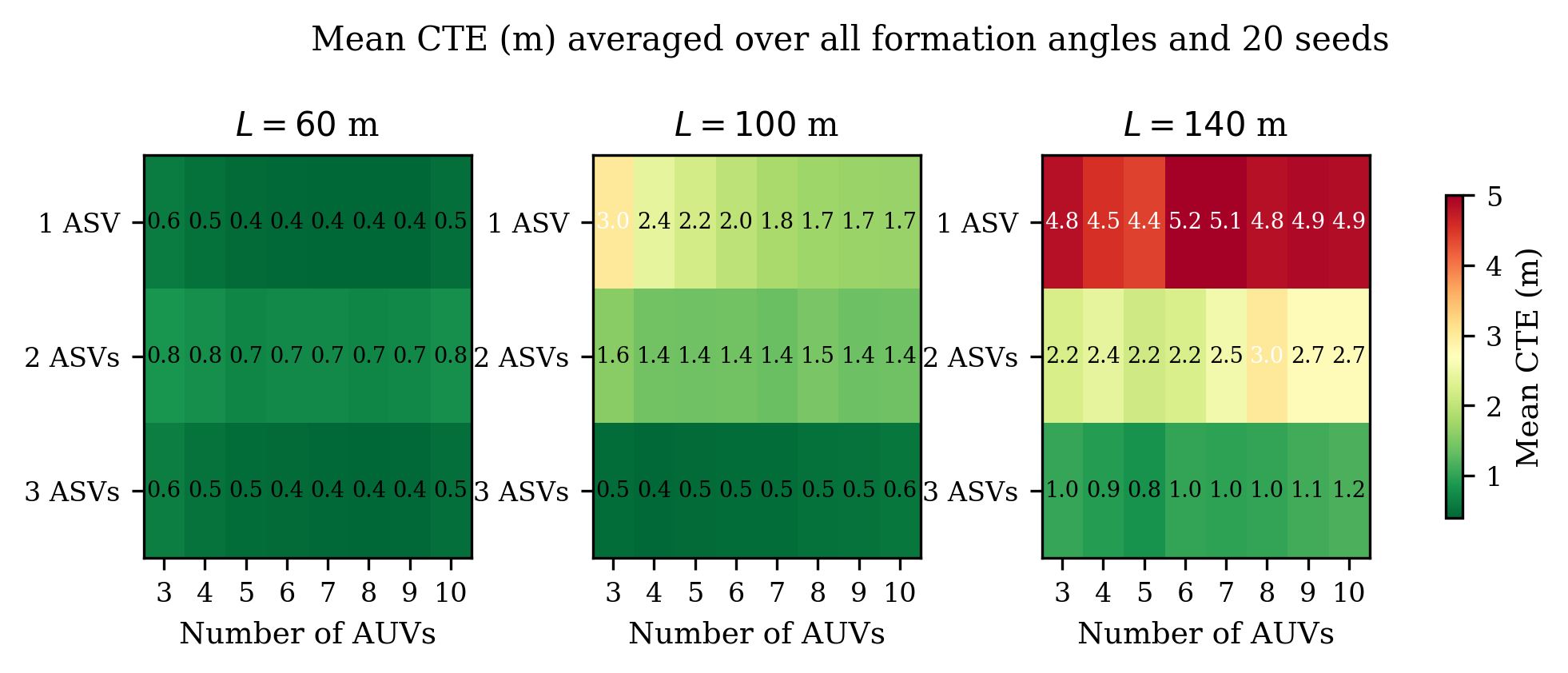}
  \caption{Mean CTE (m) across all formation angles and $20$ seeds.
  Columns: survey side lengths $60$, $100$, $140$~m. Rows: ASV count;
  horizontal axis: AUV count. Green $\leq 1$~m (well-localized);
  red $> 4$~m (navigation failure).}
  \label{fig:heatmap}
\end{figure}

 \begin{figure*}[htbp]
  \centering
  \includegraphics[width=1.6\columnwidth]{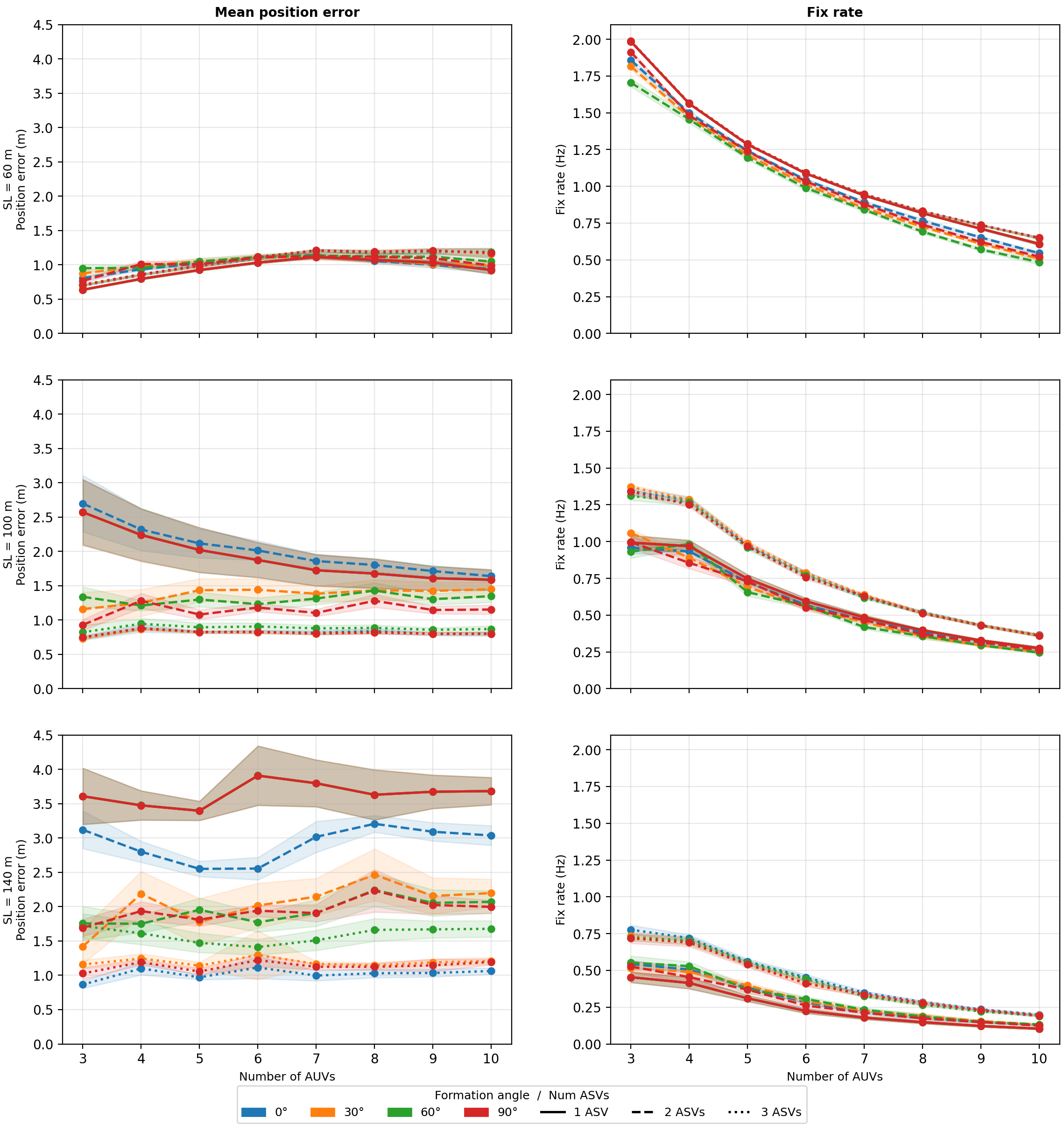}
  \caption{Distance-normalised configuration sweep ($20$ seeds, mean $\pm
  1\sigma$). Left column: mean position error. Right column: applied fix
  rate. Rows: $L = 60$, $100$, $140$~m. Colors denote formation angle;
  line styles: solid $= 1$~ASV, dashed $= 2$~ASVs, dotted $= 3$~ASVs.
  The sharp transition from solid to dotted lines at $L = 140$~m
  illustrates the binary nature of the coverage threshold.}
  \label{fig:sweep_distance_norm}
\end{figure*} 
Fig.~\ref{fig:heatmap} summarises the mean CTE across the full parameter space. 
From these results, three design principles can be inferred.
The survey scale is the dominant performance factor. At $L = 60$~m, all
configurations achieve a mean CTE below $0.8$~m regardless of the ASV count or team
size: the survey fits within the HF footprint of a single-ASV and coverage is complete.
At $L = 100$~m, single-ASV CTE grows to $1.7$--$3.0$~m, while three ASVs hold
below $0.6$~m across all team sizes. At $L = 140$~m, every single-ASV configuration
exceeds $4.4$~m universally; three-ASV formations contain this to $0.8$--$1.2$~m, a
$4$--$5\times$ reduction. This confirms the analytical coverage bound of
Section~\ref{sec:formation}: once the survey corner-to-origin distance exceeds
$R_\mathrm{HF}$, a single centroid ASV cannot provide full coverage, and additional
surface anchors are architecturally necessary rather than merely beneficial.
 
The benefit of additional ASVs is concentrated in the 1-to-3 transition. Adding a second ASV at $L = 140$~m reduces mean CTE from ${\sim}4.8$~m to
${\sim}2.5$~m. This improvement is real but angle-dependent: as the case study
(Section~\ref{sec:case_study}) showed, certain two-ASV angles still produce coverage
shadows. The third ASV provides the critical transition to sub-$1.2$~m CTE by
eliminating the remaining angle-dependent dead zones through triangular footprint
interleaving. At $L = 100$~m, adding a third ASV reduces the CTE by a factor of $3$ to $6$ 
relative to a single ASV across all AUV counts.
 
The coverage fraction governs navigation quality, not the fix rate alone. As the AUV
count increases from $3$ to $10$, the fix rate per-AUV decreases monotonically
because the TDMA budget is shared between a larger fleet. This fix-rate decay does not, 
however, translate uniformly to CTE growth. At $L = 60$~m, CTE remains flat despite
a $3\times$ fix-rate reduction because coverage is complete: all fixes that could be
delivered are delivered. At $L = 140$~m with a single ASV, CTE is already saturated
by coverage loss and is insensitive to further fix-rate changes, because the missing
fixes are not a scheduling artefact but a geometric impossibility. 
This confirms that navigation quality is governed primarily by acoustic coverage
fraction rather than fix rate, and that BIND-USBL's formation geometry is designed
to address exactly this binding constraint. 
 
Fig.~\ref{fig:sweep_distance_norm} provides the angle-resolved breakdown, showing
the mean position error and applied fix rate as a function of AUV count for each
survey scale and ASV count.  Three features are worth noting beyond the
aggregate trends of Fig.~\ref{fig:heatmap}. First, at
$L = 60$~m (top row), all curves collapse regardless of ASV
count, formation angle, or team size, confirming that the
$60$~m domain is coverage-saturated and performance is
limited only by the TDMA servicing rate. Second, at
$L = 100$~m and $140$~m, the $1$-ASV curves (solid lines)
separate sharply from the $2$- and $3$-ASV curves (dashed
and dotted), with the gap widening as AUV count decreases
--- fewer AUVs mean longer per-vehicle trajectories that
extend further into the coverage-deficient extremes of the
survey area. Third, the fix-rate curves (right column) are
nearly identical across all ASV counts at a given AUV count,
indicating that fix rate is governed by TDMA scheduling
rather than formation geometry. The divergence in position
error despite similar fix rates reinforces the finding that
coverage fraction, not servicing rate, is the binding
constraint on navigation quality at larger survey scales.\par
Taken together, the sweep results support the central claim of
BIND-USBL and make it operationally concrete. Navigation quality in heterogeneous ASV--AUV teams is governed by the
joint interaction of acoustic coverage, formation geometry, and mission scale. At
$L = 60$~m, a single ASV and the TDMA scheduler suffice; at $100$~m and beyond,
the formation design becomes the primary performance lever. The heatmap of
Fig.~\ref{fig:heatmap} provides a practical lookup table for mission planners:
given survey dimensions and fleet composition, it specifies the minimum ASV team
size needed to achieve sub-metre navigation accuracy.

\section{Conclusion} \label{con}

This paper presented BIND-USBL, a cooperative localization framework for heterogeneous ASV--AUV teams in GPS-denied underwater environments. The main result is that navigation performance is governed less by the nominal accuracy of individual USBL fixes than by their spatial and temporal availability over the mission area. A multi-ASV merged-fix scheme averages independent USBL estimates
from all surface vehicles within acoustic range, reducing single-transceiver
geometry sensitivity. With TDMA scheduling protocol and spatial-reuse extension  based on greedy graph colouring of the AUV conflict graph.  Further, a pipelined HF/MF protocol decouples uplink and downlink timing so that MF broadcasts proceed concurrently with subsequent HF uplink slots, keeping end-to-end fix latency below $0.57$\,s across all tested configurations. Simulations in HoloOcean were conducted with 3–10 AUVs and 1–3 ASVs under different formation angles and survey sizes. The results show that BIND-USBL maintains low localization error even as the mission scale increases. With three ASVs, the mean fused position error remains between 0.7–1.2 m for smaller survey areas and about 1.0–1.2 m for larger ones. In contrast, a single-ASV setup produces much higher errors, reaching about 5 m in larger survey regions. Among all tested formations for 3 ASVs, the alignment angle $\alpha_0 = 0^\circ$ gives the best performance, achieving the highest acoustic coverage and the lowest tracking error. Poorer formations, such as $\alpha_0 = 90^\circ$, significantly reduce coverage and increase localization error. As the number of AUVs increases, the localization update rate decreases because more vehicles must share the acoustic channel under TDMA scheduling.

Future work will focus on field validation, improved acoustic channel modelling, and adaptive ASV placement strategies for larger and more dynamic teams. 
Our results point toward principled design rules for heterogeneous marine robot teams, in which localization performance depends not only on onboard navigation but also on the deliberate design of cooperative surface support. This is especially relevant for persistent and large-area missions, where distributed surface support may become a key enabler of scalable underwater operations.

\section{Acknowledgment}
PK and HH acknowledge support from DFG through Germany's
Excellence Strategy--EXC 2117--422037984 and the Centre for
the Advanced Study of Collective Behaviour (CASCB),
University of Konstanz, Germany. PK and RM acknowledge the
use of ChatGPT and Claude for refining academic language and
grammar; neither tool was used to generate scientific content.

\balance
\bibliographystyle{IEEEtran}
\bibliography{biblio}
\end{document}